\documentclass[10pt,twocolumn,letterpaper]{article}

\usepackage[pagenumbers]{cvpr} 

\usepackage{graphicx}
\usepackage{amsmath}
\usepackage{amssymb}
\usepackage{siunitx}
\usepackage{booktabs}
\usepackage{CJK}
\newcommand{\myparagraph}[1]{\vspace{0.1em}\noindent\textbf{#1}}
\usepackage{comment}

\usepackage[pagebackref,breaklinks,colorlinks]{hyperref}

\usepackage[capitalize]{cleveref}
\crefname{section}{Sec.}{Secs.}
\Crefname{section}{Section}{Sections}
\Crefname{table}{Table}{Tables}
\crefname{table}{Tab.}{Tabs.}

\begin{document}

\newcommand{\aka}{a.k.a.}

\newcommand{\network}{\mathcal{F}}
\newcommand{\eqcomma}{\ \ ,}
\newcommand{\eqstop}{\ \ .}

\newcommand{\surfacepoint}{\mathbf{X}}
\newcommand{\surfacepointsample}{\surfacepoint_{inter}}
\newcommand{\imagepoint}{\mathbf{p}}
\newcommand{\projectionfunc}{\pi}
\newcommand{\pifufeatur}{f}
\newcommand{\pifufeaturglobal}{\pifufeatur^g}
\newcommand{\pifufeaturlocal}{\pifufeatur^l}
\newcommand{\pifuoccfunc}{\mathbb{F}_{occ}}
\newcommand{\pifuocc}{o}

\newcommand{\pifucolorfunc}{\mathbb{F}_{col.}}
\newcommand{\pifufeaturcol}{\pifufeatur^c}
\newcommand{\pifucolor}{c}

\newcommand{\feature}{f}
\newcommand{\featurexy}{\feature_{xy}}
\newcommand{\featurexz}{\feature_{xz}}
\newcommand{\featureyz}{\feature_{yz}}

\newcommand{\sdf}{s}
\newcommand{\sdffunc}{\mathbb{F}_{sdf}}

\newcommand{\camerapose}{\mathbf{v}}

\newcommand{\density}{\delta}

\newcommand{\origin}{\mathbf{o}}
\newcommand{\ray}{\mathbf{r}}
\newcommand{\transmittance}{T}
\newcommand{\dir}{\mathbf{d}}

\newcommand{\normalmap}{\mathbf{N}}
\newcommand{\albedomap}{\mathcal{A}}
\newcommand{\materialmap}{\mathcal{M}}
\newcommand{\depthmap}{\mathcal{D}}

\title{HumanGen: Generating Human Radiance Fields with Explicit Priors}

\author{Suyi Jiang\textsuperscript{1} \;\, Haoran Jiang\textsuperscript{1} \;\, Ziyu Wang\textsuperscript{1} \;\, Haimin Luo\textsuperscript{1} \;\, Wenzheng Chen\textsuperscript{2} \\
\;\, Lan Xu\textsuperscript{1,3}} 

\makeatletter
\let\@oldmaketitle\@maketitle
\renewcommand{\@maketitle}{
	\@oldmaketitle
	\centering
	\vspace{-8mm}
	{\large \textsuperscript{1}ShanghaiTech University}\quad \quad
	{\large \textsuperscript{2}University of Toronto}\quad \quad
	{\large \textsuperscript{3}Shanghai Engineering Research Center of Intelligent Vision and Imaging}
    
	\vspace{8mm}
}
\makeatother

\maketitle

\newcommand{\itemnudege}{\vspace{-.05in}}
\newcommand{\eqnbreak}{\par\vspace{-1.5\baselineskip}}
\newcommand{\vnudge}{\vspace{-.15in}}

\definecolor{Red}{cmyk}{0,1,1,0}
\definecolor{Green}{cmyk}{1,0,1,0}
\definecolor{Cyan}{cmyk}{1,0,0,0}
\definecolor{Purple}{cmyk}{0.45,0.86,0,0}
\definecolor{Rosolic}{cmyk}{0.00,1.00,0.50,0}
\definecolor{Blue}{cmyk}{1.00,1.00,0.00,0}
\definecolor{BlueViolet}{cmyk}{0.86,0.91,0,0.04}
\definecolor{NavyBlue}{cmyk}{0.94,0.54,0,0}

\newcommand{\other}[1]{{\color{Rosolic}  Other: #1}}
\newcommand{\su}[1]{{\color{Green}  su: #1}}
\newcommand{\guo}[1]{{\color{Red} \bf \em guo: #1}}
\newcommand{\xu}[1]{{\color{Blue} xu: #1}}
\newcommand{\jiang}[1]{{\color{BlueViolet} jiang: #1}}
\newcommand{\sun}[1]{{\color{Purple} sun: #1}}
\newcommand{\wu}[1]{{\color{Purple} wu: #1}}
\newcommand{\WaitModified}[1]{{\color{pink} sun: #1}}

\newcommand{\xuNote}[1]{{\color{red} xu: #1}}
\newcommand{\wuNote}[1]{{\color{red} wu: #1}}
\newcommand{\sunNote}[1]{{\color{red} sun: #1}}

\newtheorem{thm}{Theorem}
\newtheorem{cor}[thm]{Corollary}
\newtheorem{lem}[thm]{Lemma}
\newtheorem{prop}[thm]{Proposition}
\newtheorem{defn}[thm]{Definition}
\newtheorem{rem}[thm]{Remark}
\let\vec=\mathbf
\let\set=\mathcal
\let\mat=\mathbf

\newcommand{\mypara}[1]{\paragraph*{#1.}}
\newcommand{\bv}[1]{\mathbf{#1}}
\newcommand{\todo}[1]{\hl{#1}}
\newcommand{\bi}[1]{\mathbf{#1}}
\begin{abstract}
Recent years have witnessed the tremendous progress of 3D GANs for generating view-consistent radiance fields with photo-realism. Yet, high-quality generation of human radiance fields remains challenging, partially due to the limited human-related priors adopted in existing methods. 
We present HumanGen, a novel 3D human generation scheme with detailed geometry and $\ang{360}$ realistic free-view rendering.
It explicitly marries the 3D human generation with various priors from the 2D generator and 3D reconstructor of humans through the design of ``anchor image''.
We introduce a hybrid feature representation using the anchor image to bridge the latent space of HumanGen with the existing 2D generator.
We then adopt a pronged design to disentangle the generation of geometry and appearance.
With the aid of the anchor image, we adapt a 3D reconstructor for fine-grained details synthesis and propose a two-stage blending scheme to boost appearance generation.
Extensive experiments demonstrate our effectiveness for state-of-the-art 3D human generation regarding geometry details, texture quality, and free-view performance. Notably, HumanGen can also incorporate various off-the-shelf 2D latent editing methods, seamlessly lifting them into 3D.

\end{abstract}

\begin{CJK}{UTF8}{gbsn}
    \section{Introduction}
We are entering an era where the boundaries of real and virtually generated worlds are dismissing. An epitome of this revolution is the recent rise of 3D-aware and photo-realistic image synthesis in the past several years~\cite{schwarz2020graf,chan2021pi,zhou2021cips,EG3D,deng2022gram,or2022stylesdf,gu2021stylenerf}, which combine 2D Generative Adversarial Networks (GANs) with neural volume rendering, like neural radiance fields (NeRFs)~\cite{nerf}. But such 3D GANs mainly focus on 3D rigid contents like human/animal faces or CAD models. The further 3D generation of us humans with photo-realism is more attractive, with numerous applications in VR/AR or visual effects.

High-quality 3D human generative models should ideally generate 3D-aware humans with the following characteristics: (1) detailed geometry, (2) photo-realistic appearance, and (3) even supporting $\ang{360}$ free-view rendering.
Yet, it remains extremely challenging, mainly due to the significantly higher diversity of human apparel and skeletal pose.
Only very recently, a few work~\cite{bergman2022gnarf,zhang2022avatargen,hong2022eva3d} explore 3D GANs for human generation by using the parametric human model like SMPL~\cite{SMPL2015} as motion priors. But such parametric human prior lacks sufficient geometry details, and the adopted neural rendering in these methods does not guarantee that meaningful 3D geometry can be generated, further leading to appearance artifacts. Besides, these 3D human generators are trained with limited human datasets that lack diversity~\cite{tsuchida2019aist} or suffer from imbalanced viewing angles (most are front views)~\cite{fu2022stylegan,liu2016deepfashion}. In a nutshell, existing methods fail to fulfill all the aforementioned three characteristics for 3D human generation.

\begin{figure}[tbp] 
        \vspace{-1ex}
	\centering 
	\includegraphics[width=1\linewidth]{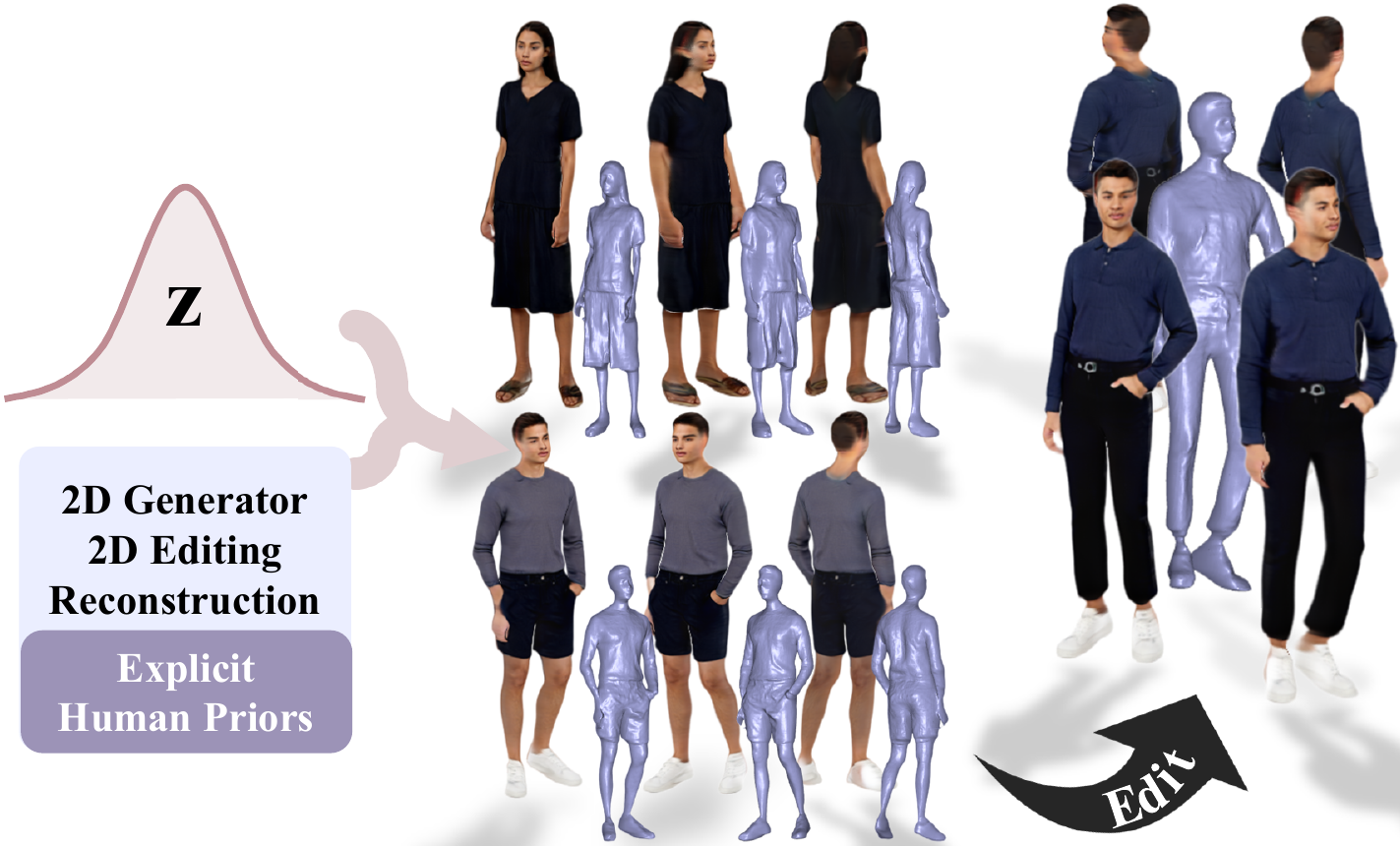}
	\vspace{-20pt} 
	\caption{The proposed HumanGen can generate 3D humans with fine-detailed geometry and appearance while seamlessly lifting various 2D latent editing tools into 3D.} 
	\label{fig:fig_1_teaser} 
	\vspace{-20pt} 
\end{figure}

We observe that 3D human generation can benefit from more explicit priors from other research domains of human modeling, except for the SMPL prior adopted in existing methods.
Specifically, with the recent large-scale dataset SHHQ~\cite{fu2022stylegan}, the 2D human generators~\cite{karras2019style,karras2020analyzing,karras2021alias} achieve more decent synthesis results than the 3D ones. And various downstream 2D editing tools are available by disentangling the latent spaces~\cite{wu2021stylespace,shen2021closed,roich2022pivotal,patashnik2021styleclip}. These abilities of 2D generation and subsequent can significantly benefit the 3D human generation if their latent spaces can be bridged. Besides, recent advances in monocular 3D human reconstruction~\cite{PIFuHD,alldieck2022photorealistic} have achieved more fine-grained geometry details than the implicit geometry proxy in current 3D human generators.
Yet, there lacks a well-designed mechanism to explicitly utilize the rich human priors from both 2D generator and 2D reconstructor for 3D human generation.

In this paper, we present \textit{HumanGen} -- a novel neural scheme to generate high-quality radiance fields for 3D humans from 2D images, as shown in Fig.~\ref{fig:fig_1_teaser}. In stark contrast with existing methods that only use SMPL, our approach explicitly utilizes richer priors from the top-tier 2D generation and 3D reconstruction schemes. As a result, our approach not only enables photo-realistic human generation with detailed geometry and $\ang{360}$ free-view ability, but also maintains the compatibility to existing off-the-shelf 2D editing toolbox based on latent disentanglement.

Our key idea in HumanGen is to organically leverage a 2D human generator and a 3D human reconstructor as explicit priors into a 3D GAN-like framework.
Specifically, we first introduce a hybrid feature representation of the generative 3D space, which consists of the tri-plane features from EG3D~\cite{EG3D} as well as a 2D photo-real human image (denoted as ``anchor image'') generated through the pre-trained 2D human generator. Note that we adopt separated StyleGAN2~\cite{karras2020analyzing} architectures to generate  both the tri-plane feature maps and the anchor image. But they share the same latent mapping network, so as to bridge and anchor the latent space of our 3D GAN to the pre-trained 2D human generator.
Then, based on such hybrid representation, we design our 3D human generator into the pronged geometry and appearance branches.
In the geometry branch, we explicitly utilize a pre-trained 3D reconstructor PIFuHD~\cite{PIFuHD} to extract pixel-aligned features from the anchor image and provide extra fine-grained geometry supervision for our HumanGen. Note that the original PIFuHD encodes geometry as an implicit occupancy field.
Thus, we propose a geometry adapting scheme to turn it into a generative version with signed distance field (SDF) output, so as to support efficient and high-resolution volume rendering with sphere tracing.  
For the appearance branch, we propose to learn an appearance field and a blending field from both the pixel-aligned and tri-plane features. Then, we adopt a two-stage blending scheme to make full use of the rich texture information in the anchor image.
For our 3D GAN training procedure, we adopt a similar training strategy like EG3D~\cite{EG3D} with pose conditioning, and  introduce additional front-view and back-view consistency supervision to enhance the generated texture details.

Besides, we observe that existing 2D human generator StyleGAN2~\cite{karras2020analyzing} trained on the large-scale SHHQ~\cite{fu2022stylegan} can potentially generate diverse human images including side-views and even back-views. Thus, 
we train our HumanGan using an augmented dataset from SHHQ by using the pre-trained 2D generator to cover $\ang{360}$ viewing angles.
Once trained, our HumanGen enables high-quality 3D human generation. As an additional benefit, it shares the same latent mapping with the 2D generated anchor image. Thus, using the anchor image, we can seamlessly upgrade off-the-shelf 2D latent editing methods into our 3D setting. We showcase various 3D effects via convenient anchor image editing. 
To summarize, our main contributions include:
\begin{itemize} 
	\setlength\itemsep{0em}
	
	\item We present a novel 3D-aware human generation scheme, with detailed geometry and $\ang{360}$ realistic free-view rendering, achieving significant superiority to state-of-the-arts.
	
	\item We propose a hybrid feature representation using an anchor image with shared latent space to bridge our 3D GAN with the existing 2D generator.
	
	\item We propose a pronged design for appearance/geometry branches, and adapt a 3D reconstructor to aid the geometry branch for  fine-grained details synthesis.
	
	\item We introduce an implicit blending field with two-stage blending strategy to generate high-quality appearance.
	
\end{itemize}

\section{Related Work} 
\noindent{\textbf{3D-aware GAN.}}
Early approaches mainly exploit explicit 3D representations for 3D-aware image synthesis, such as textured mesh~\cite{oechsle2019texture, Liao2020CVPR, szabo2019unsupervised}, and voxels~\cite{nguyen2019hologan, nguyen2020blockgan, henzler2019platonicgan, zhou2021cips, gadelha20173d, henzler2019escaping}.
These works commonly suffer from low model expressiveness or high memory footprint.
The recent NeRF~\cite{nerf} has exerted tremendous momentum towards view-consistent 3D content generation~\cite{schwarz2020graf, chan2021pi, gu2021stylenerf, or2022stylesdf, Niemeyer2020GIRAFFE, xue2022giraffehd, EG3D, deng2022gram, xiang2022gram,wang2022generative,zhao2022generative, jain2021dreamfields, hao2021gancraft, chen2022sofgan, zhou2021cips}.
However, low-resolution 3D volume generation is adopted~\cite{schwarz2020graf, chan2021pi} and compensated with 2D upsampling layers~\cite{gu2021stylenerf, or2022stylesdf, Niemeyer2020GIRAFFE, xue2022giraffehd, EG3D},  and limits the 3D view consistency and geometry detail. Hybrid representations, e.g., MPIs~\cite{deng2022gram, xiang2022gram, zhao2022generative} partially address this issue but limit the visible range.
Compared to these methods mainly focusing on the face or synthetic data~\cite{hong2022avatarclip}, quite a few recent works~\cite{EG3D, hong2022eva3d, zhang2022avatargen, bergman2022gnarf} explore 3D human generation using only 2D images.
They tend to utilize SMPL~\cite{SMPL2015} pose prior~\cite{zhang2022avatargen, bergman2022gnarf} and coarse shape prior~\cite{hong2022eva3d} to address the challenging human geometry diversity. 
However, such priors without geometry details cannot guarantee such generators generate faithful 3D geometry. Besides, the lack of view-balanced human datasets further limits their ability to around-view rendering. 
In contrast, our approach achieves generating high-quality freely-renderable human radiance fields with detailed geometry.

\begin{figure*}[t] 
    \vspace{-4ex}
	\begin{center} 
		\includegraphics[width=\linewidth]{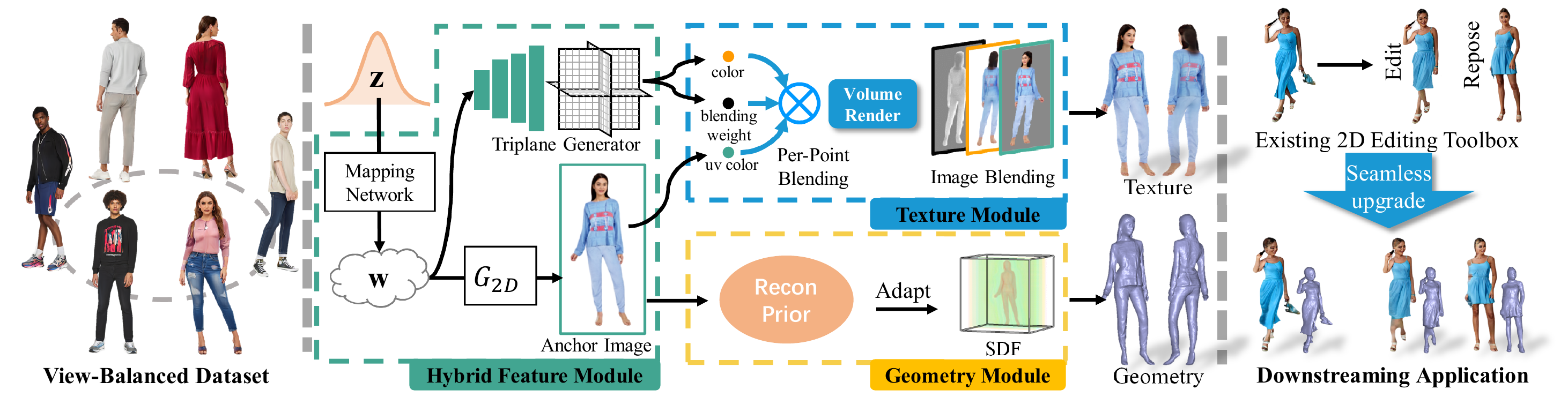} 
	\end{center} 
    \vspace{-20pt}
    \caption{Our approach consists of three modules. The hybrid feature module includes anchor image and tri-plane feature generation. The geometry module includes reconstruction prior and SDF adaptation (Sec.~\ref{sec:geometry}). Texture module includes sphere tracing based volume rendering, texture and blending weight fields, and two-stage blending (Sec.~\ref{sec:texture})} 
	\label{fig:fig_2_overview} 
	\vspace{-10pt}
\end{figure*} 

\noindent{\textbf{2D Human Image Generation.}}
A large part of 2D image generation works fall in conditional image generation~\cite{pix2pix2017, wang2018pix2pixHD, wang2018vid2vid, park2019SPADE, zhu2020sean} and achieves a photo-realistic level. As to human image synthesis, a line of work focuses on conditioning the generator with semantic map~\cite{esser2018variational, zhang20223d, jiang2022text2human}, pose~\cite{sarkar2021humangan, albahar2021pose}, texture~\cite{albahar2021pose, sarkar2021style, grigorev2021stylepeople} or even text~\cite{jiang2022text2human, hong2022avatarclip}. 
With the recent large-scale dataset SHHQ~\cite{fu2022stylegan}, the unconditional image generators~\cite{karras2020analyzing, karras2021alias, karras2019style, fu2022stylegan} achieve synthesizing appealing enough human images that are better than concurrent 3D ones.
In the meanwhile, various downstream editing techniques over faces are then extended to full-scale human editing~\cite{wu2021stylespace,shen2021closed,roich2022pivotal,patashnik2021styleclip, shen2020interfacegan}.
Rather than modeling a separate latent space for 3D human generator, we leverage the rich human appearance priors provided by 2D human generator by sharing the same latent space of a pre-trained 2D stylegan, and by the way, upgrade the editing toolchain to 3D era. 

\noindent{\textbf{Neural Human Modeling.}}
Rather than classical multi-view stereo methods that require complicated hardware, recent implicit occupancy field-based neural methods~\cite{PIFU_2019ICCV, PIFuHD, huang2020arch, he2021arch++} achieve reconstructing detailed human geometry with sparse or even one camera. 
The following works~\cite{suo2021neuralhumanfvv, MonoPort, jiang2022neuralhofusion} further reveal the effectiveness of the neural occupancy field in the real-time textured dynamic human reconstruction, equipped with a neural texture blending scheme. 
Embracing the developing of NeRF techniques~\cite{mildenhall2020nerf, wang2021ibrnet,chen2021mvsnerf, meng2021gnerf, 9466273, yu2021plenoctrees, Niemeyer2021Regnerf, mueller2022instant, verbin2022ref, chen2022tensorf, STNeRF_SIGGRAPH2021, wang2022fourier, 10.1145/3528223.3530086}, the human shape prior augmented NeRFs achieve modeling realistic human bodies~\cite{liu2021neuralActor, zhao2022humannerf, kwon2021neural, noguchi2021neural, peng2021neural}, learning animatable avatars~\cite{peng2021animatable, li2022tava, wang2022arah} and generalizing across different persons~\cite{zhao2022humannerf, wang2021ibutter, kwon2021neural} from temporal data.
However, such techniques can only build human models from actually captured data, i.e., images and videos, and cannot generate novel individuals and appearances. 
In contrast, we learn a 3D human generator from only 2D human images and largely alleviate the cost of producing high-fidelity virtual humans.

\section{Overview}
By leveraging rich priors from 2D generation and 3D reconstruction models,
HumanGen enables delicate 3D human generation with high-quality geometry details and photo-realistic textures. 
We achieve this by employing a 2D generator to synthesize an ``anchor image'' with exquisite textures, then lifting it to 3D space and utilizing a 3D reconstructor to enhance geometry details.
We briefly discuss each step below, and provide detailed explanation in Sec.~\ref{sec:algorithm}. 
%

\noindent{\bf Hybrid Feature.}
As illustrated in Fig.~\ref{fig:fig_2_overview}, 
HumanGen first employs a pretrained 2D generator $G_{2D}$ to map gaussian noise $z$ to $w$ latent space and produce an $1024\times512$ anchor image.
To lift the anchor image to 3D space, we further utilize another 3D generator (EG3D~\cite{EG3D}), synthesizing a tri-plane from $w$ to complete the missing information. 
The tri-plane is composed of three feature planes $F_{xy}$, $F_{xz}$, $F_{yz}$ which align $xy$, $xz$ and $yz$ axes, respectively. 
We then align the anchor image with $F_{xy}$ to guide the tri-pane to synthesize consistent 3D information as the anchor image.

\noindent{\bf Geometry Generation.} 
Prior works typically learn geometry in low-resolution 3D space~\cite{EG3D,gu2021stylenerf}, which produce over-smoothed human shapes. 
In contrast, we utilize priors in 3D reconstruction models~\cite{PIFuHD} to enhance more detailed shape generation.
Specifically, HumanGen also represents geometry with signed distance field (SDF). 
We employ pixel-aligned global and local feature as well as occupancy field in PIFuHD~\cite{PIFuHD} as guidance to regress SDF values and can therefore synthesize  more detailed human geometry.

\noindent{\bf Texture Generation and Blending.} 
HumanGen utilizes texture fields to decide RGB values at each 3D point and applies volume rendering to synthesize images. However, both reconstruction and generation priors help improve the texture quality. 
The former allows HumanGen to directly generate high-res images without any superresolution~\cite{gu2021stylenerf,EG3D}.
Specifically, the SDF field adapted from reconstruction prior helps restrict the sampling regions, enabling a more efficient volume rendering to synthesize $512\times256$ images directly.
We further incorporate the rich details from the anchor image to enhance the quality of rendered textures.
We learn a blending weight field and propose a two-stage blending scheme to merge the anchor image and synthesized texture.
We achieve view-consistent, photo-realistic texture generation, which we show later.

\section{Method}\label{sec:algorithm} 

\subsection{Geometry Generation} \label{sec:geometry}

Given a sampled anchor image, HumanGen first lifts it into 3D geometry.
It incorporates reconstruction priors in the lifting to synthesize fine geometry details. 
We first discuss how to extract priors from the reconstruction model~\cite{PIFuHD}, 
then describe the way to utilize them in 3D generation.

\noindent{\bf Reconstruction Priors.}
We choose PIFuHD~\cite{PIFuHD} to provide reconstruction prior, which can faithfully reconstruct fine human geometry with plausible details like hairs or wrinkles from single-view images.
Given a 3D point $\surfacepoint \in \mathbb{R}^3$, PIFuHD projects it on the image and applies a global and a local feature extractor to obtain the corresponding pixel-aligned features$(\pifufeaturglobal, \pifufeaturlocal)$. 
It further employs an implicit function $\pifuoccfunc: (\surfacepoint, \pifufeaturglobal, \pifufeaturlocal) \mapsto \pifuocc$ to map $\surfacepoint$ to an occupancy $\pifuocc \in [0, 1]$. 
We therefore collect all the features $\pifufeatur=(\pifufeaturglobal, \pifufeaturlocal)$ and the occupancy value $\pifuocc$  
as strong priors and apply them in the geometry generation.

\noindent{\bf Prior-guided SDF Adaptation.} 

We choose the SDF field to represent geometry, which has revealed better surface modelling~\cite{volsdf,wang2021neus,or2022stylesdf}.
Besides, SDF representation further allows us to derive an efficient sphere tracing to generate high-res images, which we detail in Sec.~\ref{sec:texture}.
We employ a four-layer MLP $\sdffunc: (\surfacepoint, \pifufeatur) \mapsto \sdf$ which predicts SDF value $\sdf$ given a sample point $\surfacepoint$ and PIFuHD feature $\pifufeatur$.
During training, we sample surface points $\surfacepoint_s$, where they should be 0.5 for occupancy 
denoted as: $\pifuocc(\surfacepoint_s)=0.5$.
We train $\sdffunc$ to predict correct SDF values ($\sdf(\surfacepoint_s)=0$) for those surface points, where $\mathcal{L}_{3D\_SDF}=\|\sdf(\surfacepoint_s)\|_2^2$.  
In addition to the common Eikonal loss $\mathcal{L}_{eik}=\|\nabla s(X)\|_2-1.0$ for regularizing SDF gradients, we also add mask loss to make geometry converge. %
Given a random view $\camerapose$, we get intersection mask $M_{\pifuocc,\camerapose}$ by ray marching from the occupancy field and alpha map $M_{\sdf,\camerapose}$ from the predicted SDF field. 
To get $M_{\sdf,\camerapose}$, we volume render following~\cite{or2022stylesdf} where density is calculated by: $\sigma(\surfacepoint) = \alpha^{-1} \text { sigmoid }(-s(\surfacepoint) / \alpha)$ and $\alpha$ here is a learnable parameter
.The mask loss is to minimize their difference, where $\mathcal{L}_{mask,\camerapose} =\|M_{\pifuocc,\camerapose}-M_{\sdf,\camerapose}\|_2^2 $.
The final geometry loss is defined as
\begin{equation}
	\mathcal{L}_{geo}= \lambda_{mask}\mathcal{L}_{mask,v} + \lambda_{3D\_SDF}\mathcal{L}_{3D\_SDF} + \lambda_{eik}\mathcal{L}_{eik}.
\label{eqn1} 
\end{equation}
Following PIFu~\cite{PIFU_2019ICCV}, we further train a color module that will be further used in later texture generation. 
We first employ a CNN to extract image features. For a 3D point $\surfacepoint$, we project it on the feature map and collect color feature $\pifufeaturcol$. 
We then train the color module $\pifucolorfunc: (\pifufeaturcol, \pifufeatur) \mapsto \pifucolor \in [0, 1]^3$, and supervise the loss between predicted color and GT sampled color from textured meshes.

\begin{figure}[t] 
        
	\begin{center} 
		\includegraphics[width=1\linewidth]{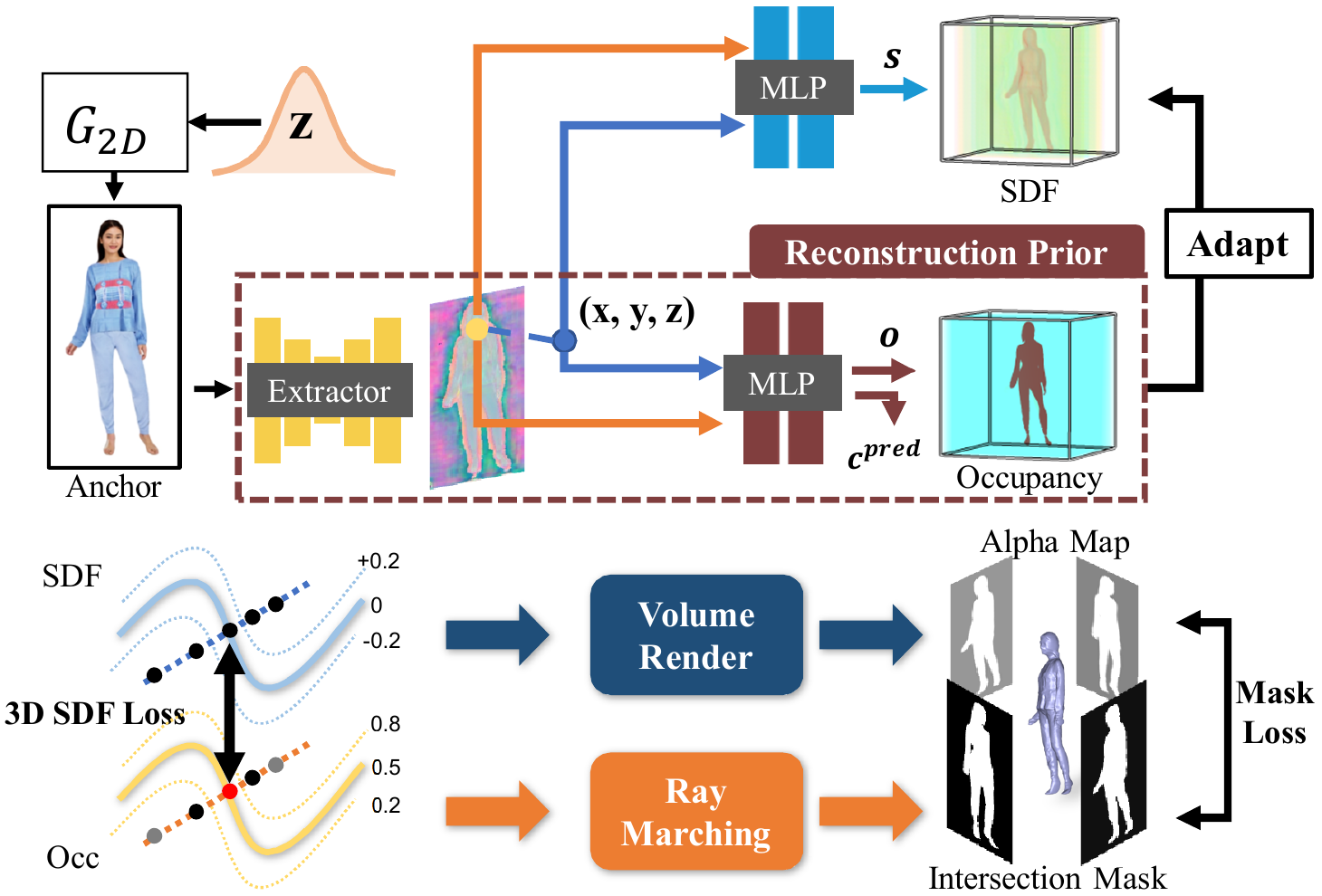}
	\end{center} 
	\vspace{-1ex}
	\caption{Illustration of reconstruction prior and our SDF adaptation scheme.(Sec.~\ref{sec:geometry})} 
	\label{fig:human_geometry_pipeline} 
	\vspace{-10pt}
\end{figure}

\begin{figure*}[t] 
        \vspace{-1ex}
	\begin{center} 
		\includegraphics[width=1\linewidth]{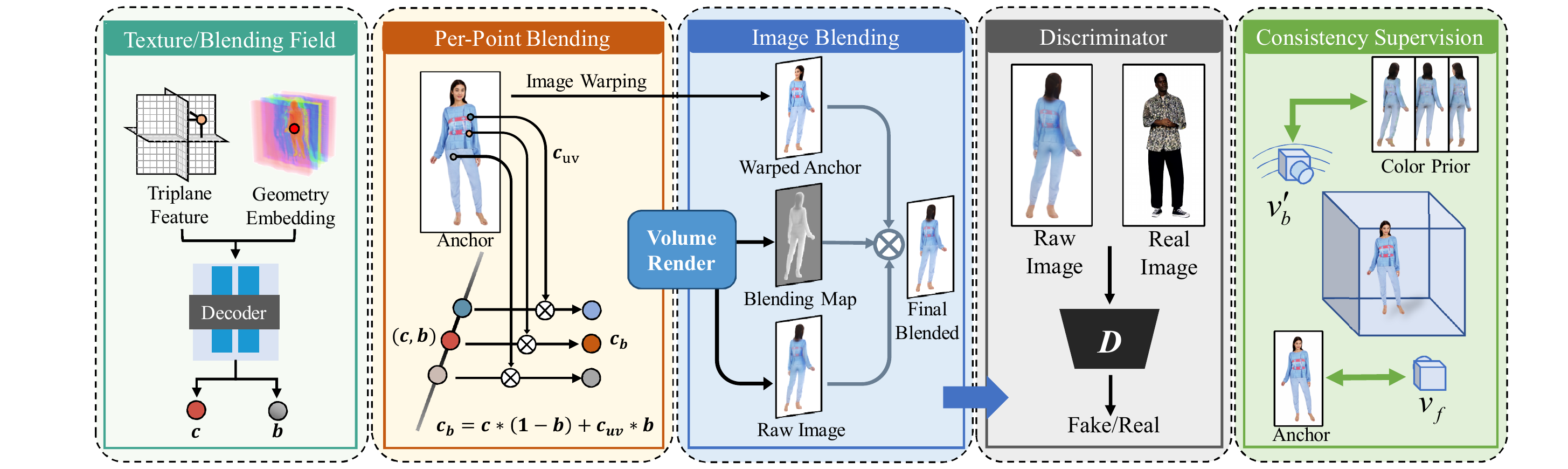}
	\end{center} 
	\vspace{-20pt}
	\caption{Illustration of texture and blending weight fields, two-stage blending and consistency supervision.(Sec.~\ref{sec:texture})} 
	\label{fig:human_texture_pipeline} 
	\vspace{-10pt}
\end{figure*} 

\subsection{Texture Generation and Blending.}   \label{sec:texture}

\noindent{\bf Volume Render with Sphere Tracing.}
As illustrated in Fig.\ref{fig:human_texture_pipeline}, geometry branch takes anchor image from $G_{2D}$ and generates the corresponding SDF field,
which allows us to perform sphere tracing to find the surface where $\sdf(\surfacepoint)=0$. Specifically, we choose an orthogonal camera model fitting with the PIFuHD setting to render. For each ray, we first query $s(\surfacepoint_0)$ at $\surfacepoint_0 = \mathbf{o} + t_{start}\mathbf{d}$ where $\mathbf{o}$ and $\mathbf{d}$ is ray origin and direction, $t_{start}$ is the pre-defined starting step. Then it iterates to query $s(\surfacepoint_n)$ at $\surfacepoint_n = \mathbf{o} + (t_{start} + \sum_{i=0}^{n-1} s(\surfacepoint_i))\mathbf{d}$, until $s(\surfacepoint_n)$ converges to 0 or iteration exceeds $n_{max}$ times. We set it to be 12 empirically. With the intersection point of each ray with geometry, we only sample 6 points uniformly around the intersection point to efficiently apply volume rendering to synthesize images. The minimal sampling number allows us to generate $512\times256$ high-res images.

\noindent{\bf Texture and Blending Weight Field.}  
With tri-plane from hybrid feature generation,
for any queried point $\surfacepoint$ in 3D space, it is projected onto each feature plane to get feature $\featurexy$, $\featurexz$ and $\featureyz$. We model a two-layer MLP following~\cite{EG3D} as an implicit decoder to decode color $c \in [0,1]^3$ and blending weight $b \in [0,1]$. To make texture generation branch geometry-aware, we also apply the PIFuHD feature $\pifufeatur$ from Sec.\ref{sec:geometry}: $f_{decoder}: (E(\featurexy, \featurexz, \featureyz), \pifufeatur) \mapsto (c, b)$. $E$ denotes the mean operation. For convenience, we use $c(\surfacepoint)$ and $b(\surfacepoint)$ to denote color and blending weight at $\surfacepoint$.

\noindent{\bf Two-Stage Blending.}
Our goal is to synthesize a high-detailed texture map. While texture field tends to produce under-detailed results, we further blend it with anchor image to enhance details.
Before volume rendering, each sample point on rays will query its SDF $s$, RGB $c$ and blending weight $b$. To incorporate information from anchor image, sample points are also projected to anchor image to fetch pixel-aligned RGB $c_{uv}(\surfacepoint)$. $c_{uv}(\surfacepoint)$ and $c(\surfacepoint)$ are then blended through $c_b(\surfacepoint) = c(\surfacepoint) \cdot (1 - b(\surfacepoint)) + c_{uv}(\surfacepoint) \cdot b(\surfacepoint)$. 

Then we render $512\times256$ raw RGB image $I_{raw}$ and blending map $I_{map}$ with volume rendering. For each ray, we get integral color $C(\mathbf{r})$ and blending weight $B(\mathbf{r})$ of ray $\mathbf{r}(t)=\mathbf{o} + t\mathbf{d}$ by sampling 6 sample points in an interval surrounding its intersection point with geometry. The sampling interval $[t_{start}, t_{end}]$ is empirically determined by convergence of $\alpha$ in Sec.\ref{sec:geometry}
\begin{equation}
\begin{split}
\begin{aligned}
C(\mathbf{r})=\int_{t_{st}}^{t_{ed}} T(t) \sigma(\mathbf{r}(t)) c_b(\mathbf{r}(t)) d t, \\
B(\mathbf{r})=\int_{t_{st}}^{t_{ed}} T(t) \sigma(\mathbf{r}(t)) b(\mathbf{r}(t)) d t, \\
T(t)=\exp \left(-\int_{t_{st}}^t \sigma(\mathbf{r}(s)) d s\right).
\end{aligned} 
\end{split}
\label{equ2} 
\end{equation}
Since points with same $xy$ coordinates blend same color from 2D image, from side view, the raw rendering result tends to have "stretching" artifact. 
To alleviate such artifact, we further utilize calculated blending map to blend rendered image $I_v$ with warped anchor image as post-processing: $I_v = I_{raw} \cdot (1 - I_{map}) + \hat{I}_{anchor} \cdot I_{map}$ where $I_v$ denotes final image from specific rendering view $v$, $\hat{I}_{anchor}$ denotes warped anchor image from frontal view to render view.


\noindent{\bf Consistency Supervision.} \label{sec:CS}
To enhance 3D consistency with the 2D anchor image, we design two consistency losses $\mathcal{L}^{CS}_{front}$ and $\mathcal{L}^{CS}_{back}$. 
Firstly, the anchor image $I$ from $G_{2D}$ is aligned with $xy$ axes, so it can naturally be used to supervise rendered image $I_{v_f}$ from frontal camera $v_f$ viewing through z-axis with 2D photometric loss $\mathcal{L}_{2D}$.
Besides, from the frontal camera $v_f$, we calculate the ray intersection points $\surfacepointsample$ with sphere tracing. The decoded color $c(\surfacepointsample)$ is supervised to be close to their pixel-aligned RGB color $c_{uv}(\surfacepointsample)$ from anchor image by 3D RGB loss. Finally, we add a regularization term $\mathcal{L}_{reg}$ on the blending weight of $\surfacepointsample$ to enhance the blending effect on the frontal side. The final loss is as $\mathcal{L}^{CS}_{front}=\lambda_{2D\_front}\mathcal{L}_{2D, v_f} + \lambda_{3D\_RGB}\mathcal{L}_{3D\_RGB} + \lambda_b\mathcal{L}_{reg,b}$, where
\begin{equation}
\begin{split}
\begin{aligned}
\mathcal{L}_{2D, v_f}&=\|I_{v_f} - I\|_2^2, \\
\mathcal{L}_{3D\_RGB}&=\|c(\surfacepointsample) - c_{uv}(\surfacepointsample)\|_2^2, \\
\mathcal{L}_{reg,b}&=\|1 - b(\surfacepointsample)\|_2^2.
\end{aligned}
\end{split}
\label{equ5} 
\end{equation}
To supervise consistency on the back view,
we use the pretrained reconstruction color prior as mentioned in Sec.~\ref{sec:geometry} to calculate consistency loss $\mathcal{L}^{CS}_{back}$. For rendering view $v_b$ opposite to $v_f$, a standard gaussian noise is added to its spherical coordinates to get random view $v^{\prime}_b$ and 2D photometric loss is calculated between volume rendered image $I_{v^{\prime}_b}$ and predicted color image $I^{pred}_{v^{\prime}_b}$ which is calculated by query $f_c(\surfacepointsample)$ on intersection points. $\mathcal{L}^{CS}_{back}$ is as:
\begin{equation}
\begin{split}
\begin{aligned}
\mathcal{L}^{CS}_{back}=\lambda_{2D\_back}\mathcal{L}_{2D,v^{\prime}_b}, \\
\mathcal{L}_{2D,v^{\prime}_b} =\|I_{v^{\prime}_b} - I^{pred}_{v^{\prime}_b}\|_2^2.
\end{aligned}
\end{split}
\label{equ6} 
\end{equation}

\subsection{Training.} \label{sec:training}

\noindent{\bf Traning Set.} 
Current 2D human image collections typically have view distribution bias~\cite{fu2022stylegan,UBCfashion}, as they focus more on taking photos in front views. Previous method~\cite{zhang2022avatargen} use sampling trick to alleviate this. 
We notice that $G_{2D}$ also generates view-biased images and produces more diverse side view and back view at a lower frequency. So we apply a human pose calibration~\cite{pymaf} on $G_{2D}$ to filter images and generate a relatively view-balanced training set of 230k images.

\begin{figure*}[htbp] 
	\begin{center} 
		\includegraphics[width=1.0\linewidth]{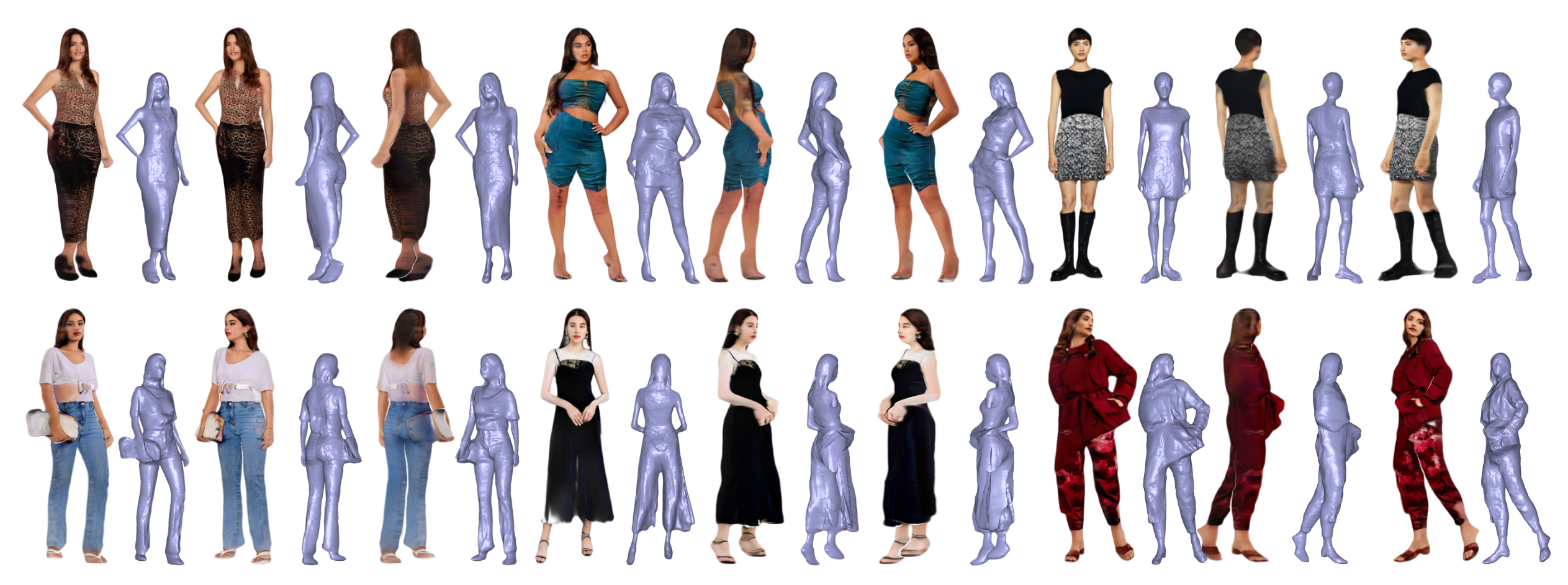} 
	\end{center} 
	\vspace{-20pt}
	\caption{The geometry and texture generation results of our HumanGen on various identities.} 
	\label{fig:fig_all}
	\vspace{-20pt}
\end{figure*}

\noindent{\bf GAN Training.} 
We further use adversarial loss to refine whole texture generation. For previous triplane-based methods, they all assume that the learned object is in a canonical state and thus condition discriminator on live state parameters like camera pose and human skeleton pose. In our setting, we assume that images from $G_{2D}$ are aligned with $xy$ axes. However, human in image is probably not facing to front, so we condition discriminator on relative pose of human. In order to render with same view distribution with dataset, for image $I$ given by $G_{2D}$, we calibrate it with PyMAF~\cite{pymaf} to get relative human root transformation $M_{human}$ and skeleton pose. For sampled camera view $v$ from dataset, it is transformed with $v_r = {M_{human}}^{-1}v$, where $v_r$ denotes actual rendering view. The GAN loss is %
\begin{equation}
\begin{split}
\begin{aligned}
\mathcal{L}_{adv}(\theta_G, \theta_D)=\boldsymbol{E}_{z \sim p_z, v \sim p_{dst}}[\log(D(G(z,v ; \theta_G)))] \\
+\boldsymbol{E}_{\I_r \sim p_{real}}[\log(-D(I_r ; \theta_D))+\lambda\|\nabla D(I_r ; \theta_D)\|_2^2].
\end{aligned}
\end{split}
\label{equ7} 
\end{equation}
HumanGen training is separated into first stage training for geometry branch and second stage training for texture branch. More details can be found in the supplementary.

    \section{Experimental Results} 

\vspace{-0.02in}

In this section, we evaluate our HumanGen on quality of generated texture and geometry. Metrics used are Frechet Inception Distance (FID)~\cite{FIDmetric} and Kernel Inception Distance (KID)~\cite{KIDmetric}. 50k generated images are used to compute scores. However, FID, KID cannot properly assess geometry quality, so we further evaluate geometry by comparing generated depth with aligned depth predicted by MiDaS~\cite{MiDaS} in masked region on 5k generated samples. 

\subsection{Comparison} 
\begin{figure}[htbp] 
	\begin{center} 
		\includegraphics[width=0.95\linewidth]{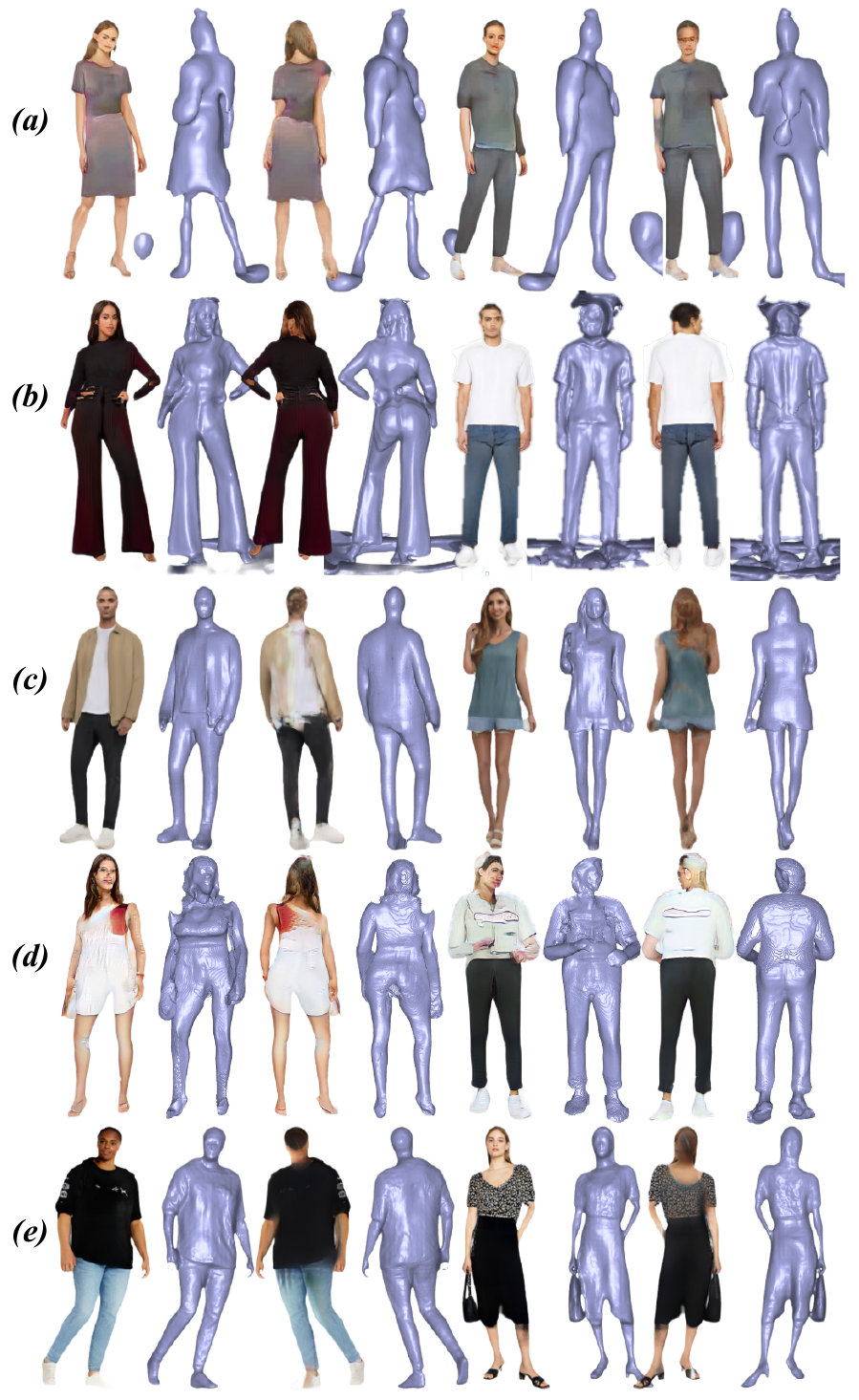}
	\end{center} 
	\vspace{-20pt}
	\caption{Qualitative comparison. (a) StyleSDF (b) EG3D (c) Stylegan-human + PIFu (d) GNARF (e) Ours}
	\label{fig:qualitative_comparison}
	\vspace{-20pt}
\end{figure}

We compare our method with state-of-the-art 3D-aware generation methods, EG3D~\cite{EG3D}, StyleSDF~\cite{or2022stylesdf}. Another compared baseline is the combination of Stylegan-human~\cite{fu2022stylegan} and PIFu~\cite{PIFU_2019ICCV}. Besides, we compare GNARF~\cite{bergman2022gnarf} implemented by ourselves with super-resolution module to achieve same resolution. As illustrated in Fig.~\ref{fig:qualitative_comparison}, StyleSDF fails to generate full-body and diverse texture and performs poor on generated geometry. EG3D generates wrong geometry and its texture appearance lacks view consistency, especially on the head region.
GNARF generates geometry with artifacts on body parts because of self-intersection problem caused by imperfect pose calibration from single-view human image.
2D Generator with PIFu generates correct full-body geometry but lacks fine geometry details and its texture tends to be blurred. While our HumanGen achieves better detailed full-body geometry as well as photo-realistic texture generation. We compare above methods on FID and KID for texture and depth for geometry. The quantitative results in Tab.~\ref{table:quatitative_comparison} demonstrate that our method achieves the best FID and depth.

\begin{table}[t]
	\begin{center}
		\centering
		\caption{Quantitative comparison of generation results.}
		\vspace{-10pt}
		\label{table:quatitative_comparison}
		\resizebox{0.45\textwidth}{!}{
			\begin{tabular}{l|cccc}
				\hline
				Method      & FID$\downarrow$ & KID$\downarrow$ & Depth$\downarrow$ \\
				\hline
				EG3D ~\cite{EG3D}\qquad\qquad & 21.33       &  \textbf{0.0110} &  0.0395   \\
				StyleSDF  ~\cite{or2022stylesdf}       &  36.69     &  0.0309  &  0.0493 \\
				2D-G~\cite{fu2022stylegan}+PIFu~\cite{PIFU_2019ICCV}\qquad\qquad &  39.20& 0.0351  &  0.0379 \\
                    GNARF\cite{bergman2022gnarf}    &     24.61 &  0.0169  &   0.0408  \\
				Ours        &  \textbf{20.97} & 0.0157 &  \textbf{0.0201}  \\
				\hline
			\end{tabular}
		}
		\vspace{-20pt}
	\end{center}
\end{table}

\subsection{Ablation Study} \label{sec:abla} 

\begin{figure}[tbp] 
	\begin{center} 
		\includegraphics[width=0.90\linewidth]{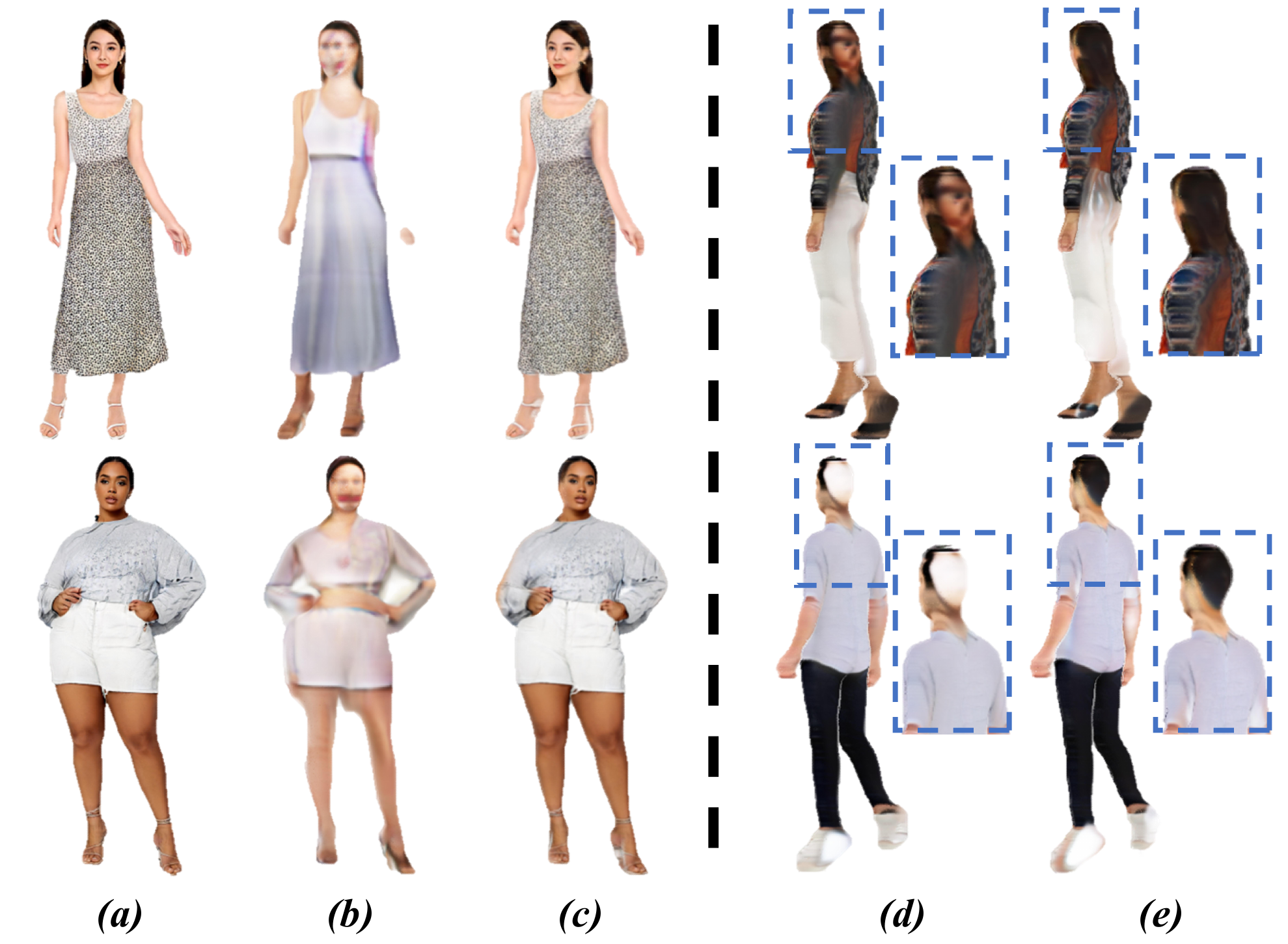} 
	\end{center} 
	\vspace{-20pt}
	\caption{Qualitative evaluation of texture generation. (a) Anchor image. (b) base; (c)\&(e) Ours  (d) w/o relative cam pose condition.} 
	\label{fig:evaluation1} 
	\vspace{-10pt}
\end{figure}

\myparagraph{Adversarial Training.}
We evaluate texture generation based on the same fixed geometry branch. In Fig.~\ref{fig:evaluation1}(b), we train a tri-plane generator to decode texture only with adversarial loss (Eqn.~\ref{equ7}) and discriminator without condition (\textbf{base}). We find it hard for the generator to maintain color or identity consistency with the anchor image and generate high-quality texture of human part.
In Fig.~\ref{fig:evaluation1}(d), without relative camera pose conditioned on discriminator, the generator tends to be confused about correct body part position.
While our method maintains high consistency with given anchor image and learns to generate full-body texture. Quantitative result in Tab.~\ref{table:quatitative_evaluation} demonstrates that our method achieves better score.

\begin{figure}[tbp] 
	\begin{center} 
		\includegraphics[width=0.90\linewidth]{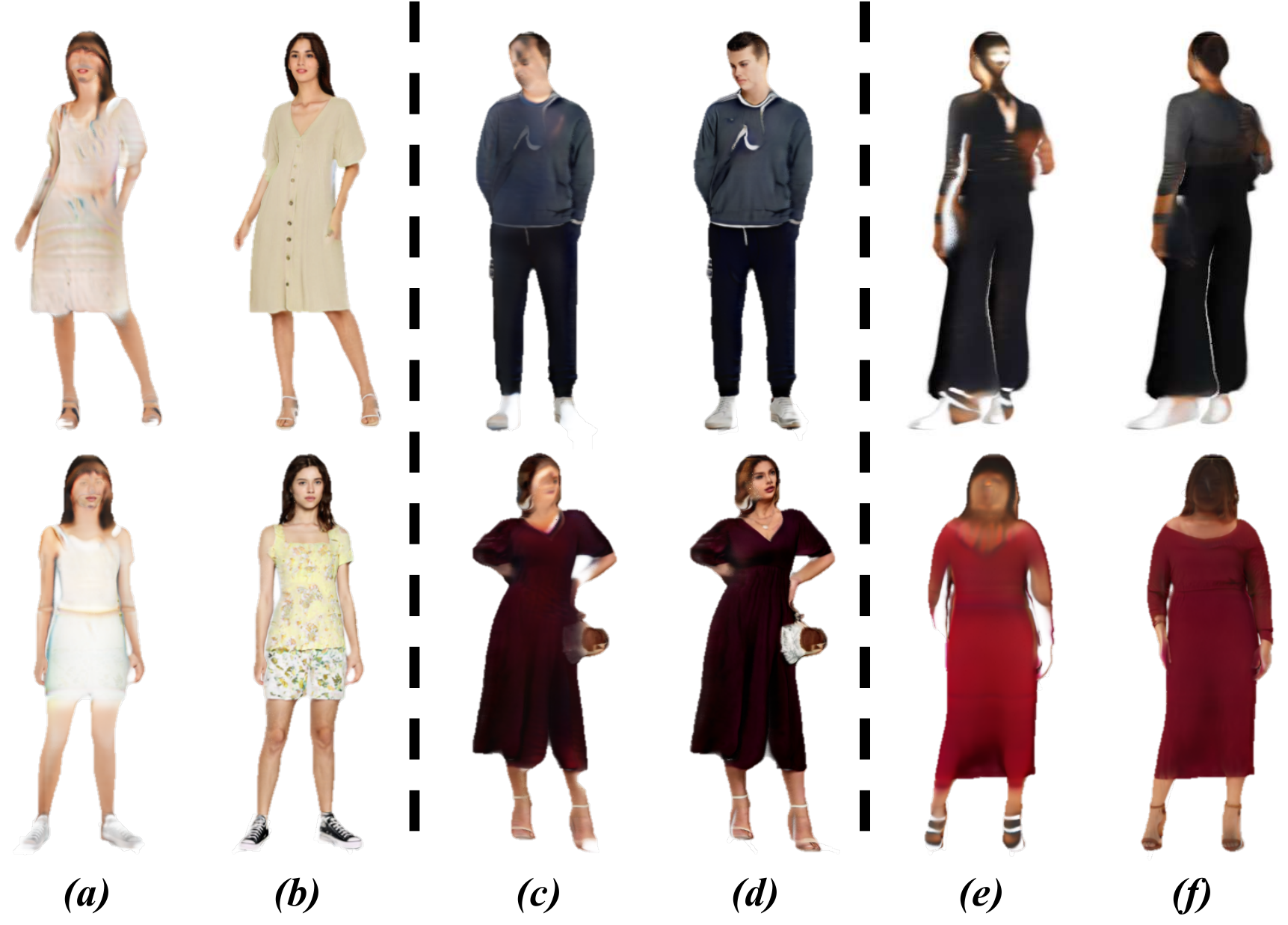} 
	\end{center} 
	\vspace{-20pt}
	\caption{Qualitative evaluation of texture generation. (a) w/o both CS. (b)\&(d)\&(f) Ours ; (c) w/o frontal CS; (e) w/o back CS.} 
	\label{fig:evaluation2} 
	\vspace{-10pt}
\end{figure}

\myparagraph{Consistency Supervision.}
Let CS denote consistency supervision described in Sec.~\ref{sec:CS}. In Fig.~\ref{fig:evaluation2} (a), \textbf{without both CS} from Eqn.~\ref{equ5} and Eqn.~\ref{equ6}, the generator is prone to change cloth type or color and is not consistent with anchor image. In (c) and (e), \textbf{without frontal CS} or \textbf{without back CS}, the texture generation on unsupervised side is prone to disobey color or identity consistency with supervised side. Our full method maintains consistency with anchor image while achieving the best self-consistency. As in Tab.~\ref{table:quatitative_evaluation}, our full method achieves better score. 

\begin{figure}[tbp] 
        \vspace{-1ex}
	\begin{center} 
		\includegraphics[width=0.90\linewidth]{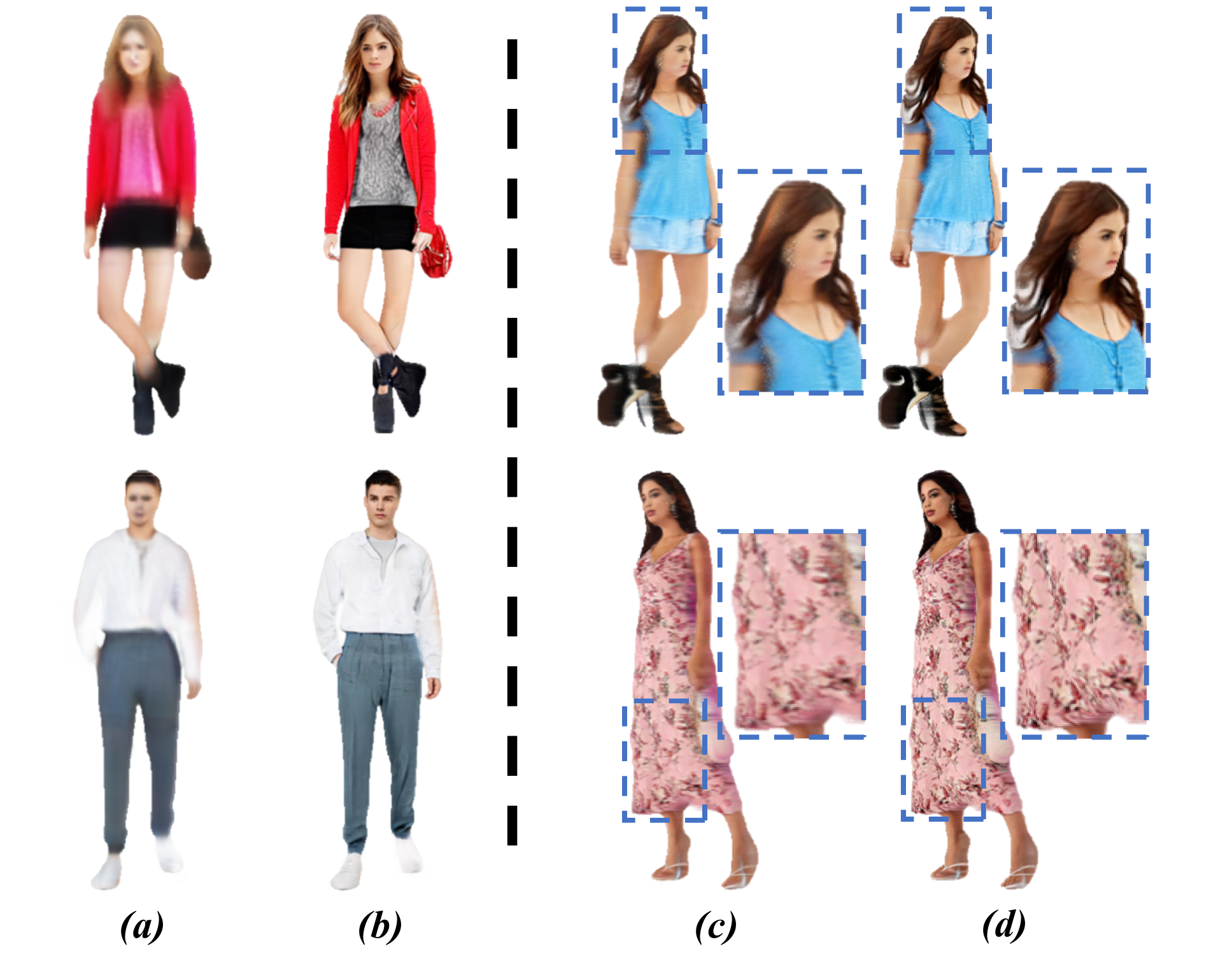} 
	\end{center} 
	\vspace{-20pt}
	\caption{Qualitative evaluation of texture generation. (a) w/o two-stage blending scheme. (b)\&(d) Ours; (c) w/o image blending.} 
	\label{fig:evaluation3} 
	\vspace{-10pt}
\end{figure}

\myparagraph{Two-Stage Blending.}
In Fig.~\ref{fig:evaluation3} (a), \textbf{without both blending scheme}, the texture generation depends completely on decoded RGB, which is prone to be blurred and cannot recover high-fidelity texture consistent with anchor image. 
The second-stage image blending depends on learned first-stage per-sample-point blending. As illustrated in (c), \textbf{without image blending} as second-stage post-processing, the texture will have "stretching" artifacts because sample points sharing the same $xy$-coordinates are blended with the same pixel color from anchor image. In our full method, per-sample-point blending enables recovering high-frequency details and image blending alleviates artifacts of first-stage blending. Quantitative results can refer to Tab.~\ref{table:quatitative_evaluation}.

\begin{figure}[tbp] 
	\begin{center} 
		\includegraphics[width=0.90\linewidth]{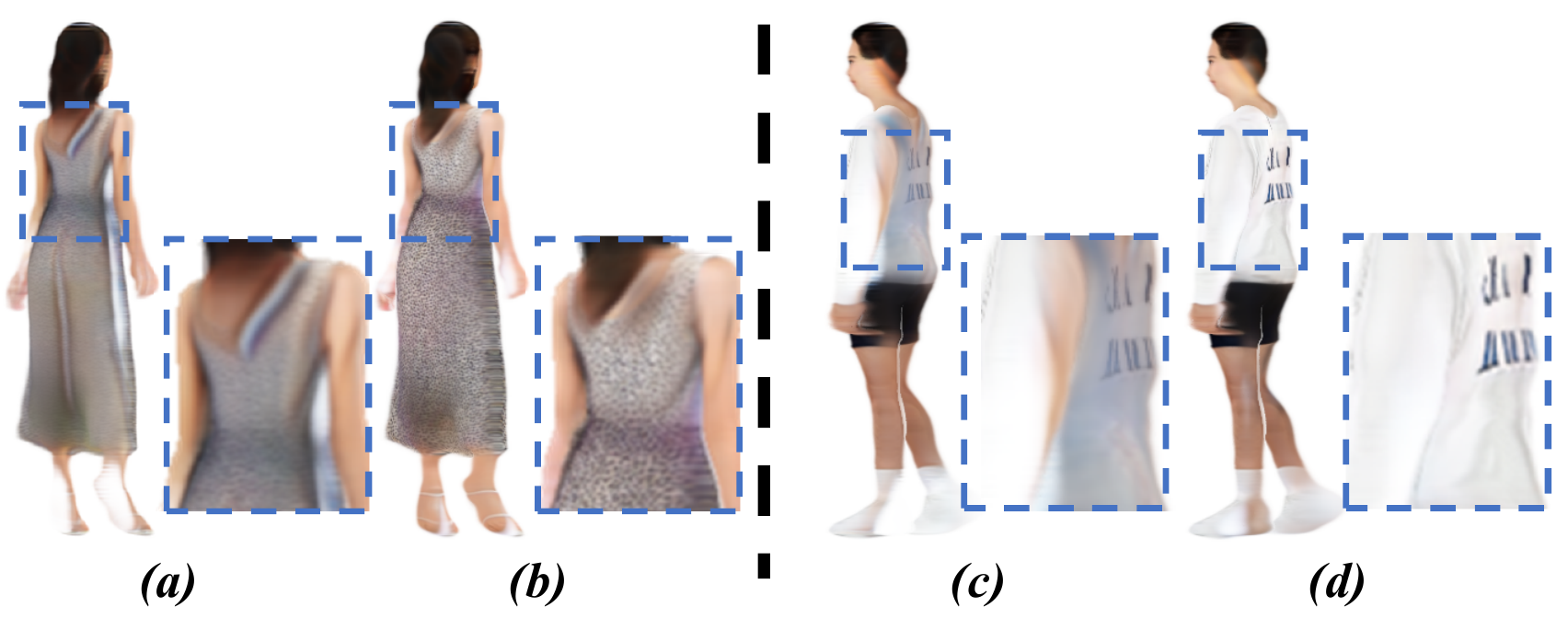} 
	\end{center} 
	\vspace{-20pt}
	\caption{Qualitative evaluation of view-distribution of training set. (a)\&(c) results on view-biased training set; (b)\&(d) Ours.} 
	\label{fig:evaluation4} 
	\vspace{-10pt}
\end{figure}

\myparagraph{View Distribution of Training Set.}
As for evaluation of view distribution of training set, we further train our full method on another synthesised view-biased training set without data from back views. As shown in Fig.~\ref{fig:evaluation4}, without adversarial loss on back \ang{180} region, the texture generation on back tends to degrade to predicted color prior from reconstruction model which is prone to be blurred and has obvious color difference with frontal texture. Quantitative results can refer to Tab.~\ref{table:quatitative_evaluation}.

\begin{figure}[tbp] 
	\begin{center} 
		\includegraphics[width=0.90\linewidth]{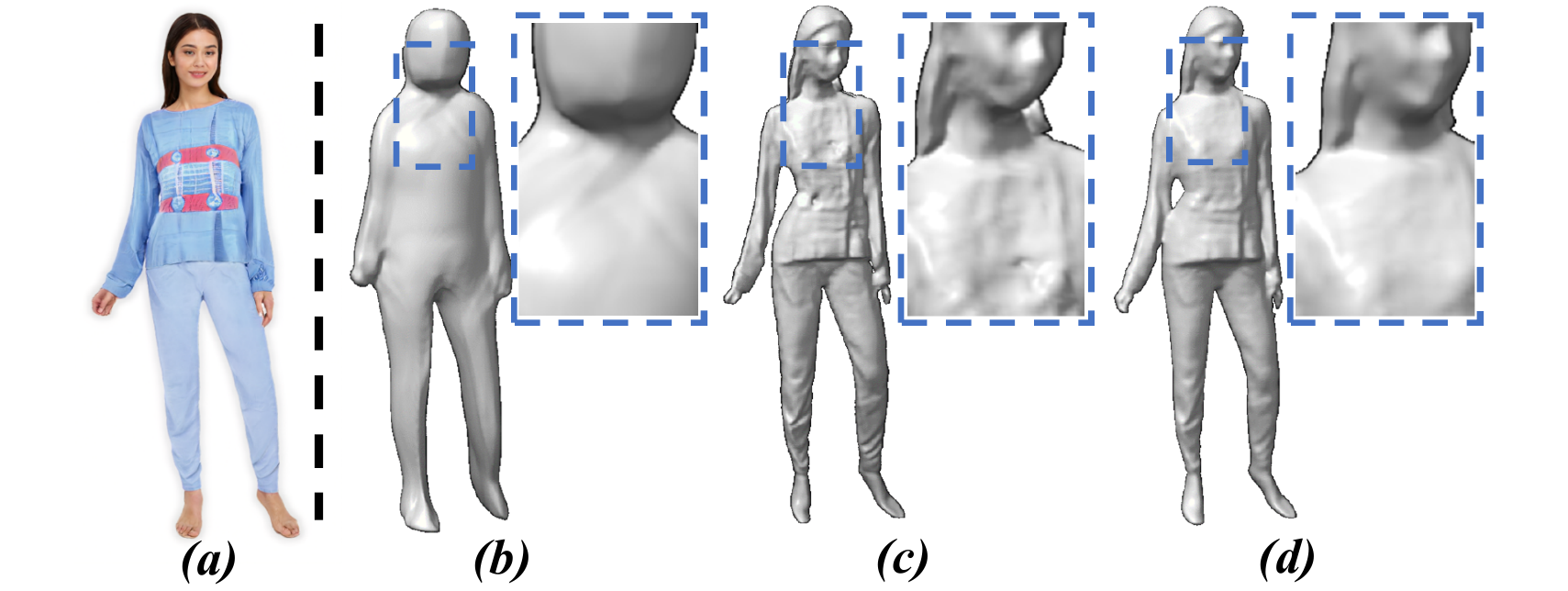} 
	\end{center} 
	\vspace{-20pt}
	\caption{Qualitative evaluation of human geometry generation. (a) Anchor images. (b) w/o PIFuHD feature; (c) w/o SDF; (d) Full.} 
	\label{fig:human_geometry} 
	\vspace{-10pt}
\end{figure}

\myparagraph{Analysis on Geometry Adaptation.}
We conduct an analysis on geometry adaptation with reconstruction prior. In Fig.~\ref{fig:human_geometry} (b), without PIFuHD~\cite{PIFuHD} feature, we directly adapt a tri-plane generator with losses from Eqn.~\ref{eqn1}, but the geometry fails to converge. In (c)\&(d), we respectively adapt an MLP to output density and SDF. Geometry in SDF representation is more smooth and has clear surface level set. As shown in Tab.~\ref{table:evaluation_geometry}, with better representation of SDF, our full method achieves the lowest difference with depth from PIFuHD.

\begin{table}[t]
    \footnotesize
	\begin{center}
		\centering
		\caption{Quantitative evaluation of texturing generation.}
		\vspace{-10pt}
		\label{table:quatitative_evaluation}
		\resizebox{0.45\textwidth}{!}{
			\begin{tabular}{l|cccc}
				\hline
				Method      & \qquad FID$\downarrow$ & KID$\downarrow$  \\
				\hline
                base                      & \qquad31.07            & 0.0251    \\
                \hline
				\hline 
                w/o both CS                & \qquad34.52           & 0.0246      \\
                w/o frontal CS             & \qquad25.75           & 0.0193      \\
                w/o back CS                & \qquad21.69           & 0.0137      \\
                \hline
			\hline 
			  w/o both blending scheme  \qquad\qquad & \qquad48.37           & 0.0436       \\
                w/o image blending         & \qquad25.94           & 0.0193       \\
                \hline
			\hline 
                full (view-biased dataset) & \qquad21.81           & 0.0161  \\
                full                       &\qquad \textbf{20.97}   & 0.0157 \\
			\end{tabular}
		}
		\vspace{-20pt}
	\end{center}
\end{table}

\begin{table}[t]
	\begin{center}
		\centering
		\caption{Quantitative evaluation of geometry generation.}
		\vspace{-10pt}
		\label{table:evaluation_geometry}
		\resizebox{0.45\textwidth}{!}{
			\begin{tabular}{l|cccc}
				\hline
				Method       
				& \qquad Depth Diff $\downarrow$ \\
				\hline
				w/o PIFuHD feat. \qquad\qquad    &  \qquad\qquad\qquad 0.0383 \qquad\qquad\qquad        \\
				w/o SDF       & \qquad\qquad\qquad 0.0183 \qquad\qquad\qquad         \\
				Full        & \qquad\qquad\qquad\textbf{0.0145}\qquad\qquad\qquad                             \\
				\hline
			\end{tabular}
		}
		\vspace{-20pt}
	\end{center}
\end{table}

\begin{figure}[tbp] 
        \vspace{-1ex}
	\begin{center} 
		\includegraphics[width=1.0\linewidth]{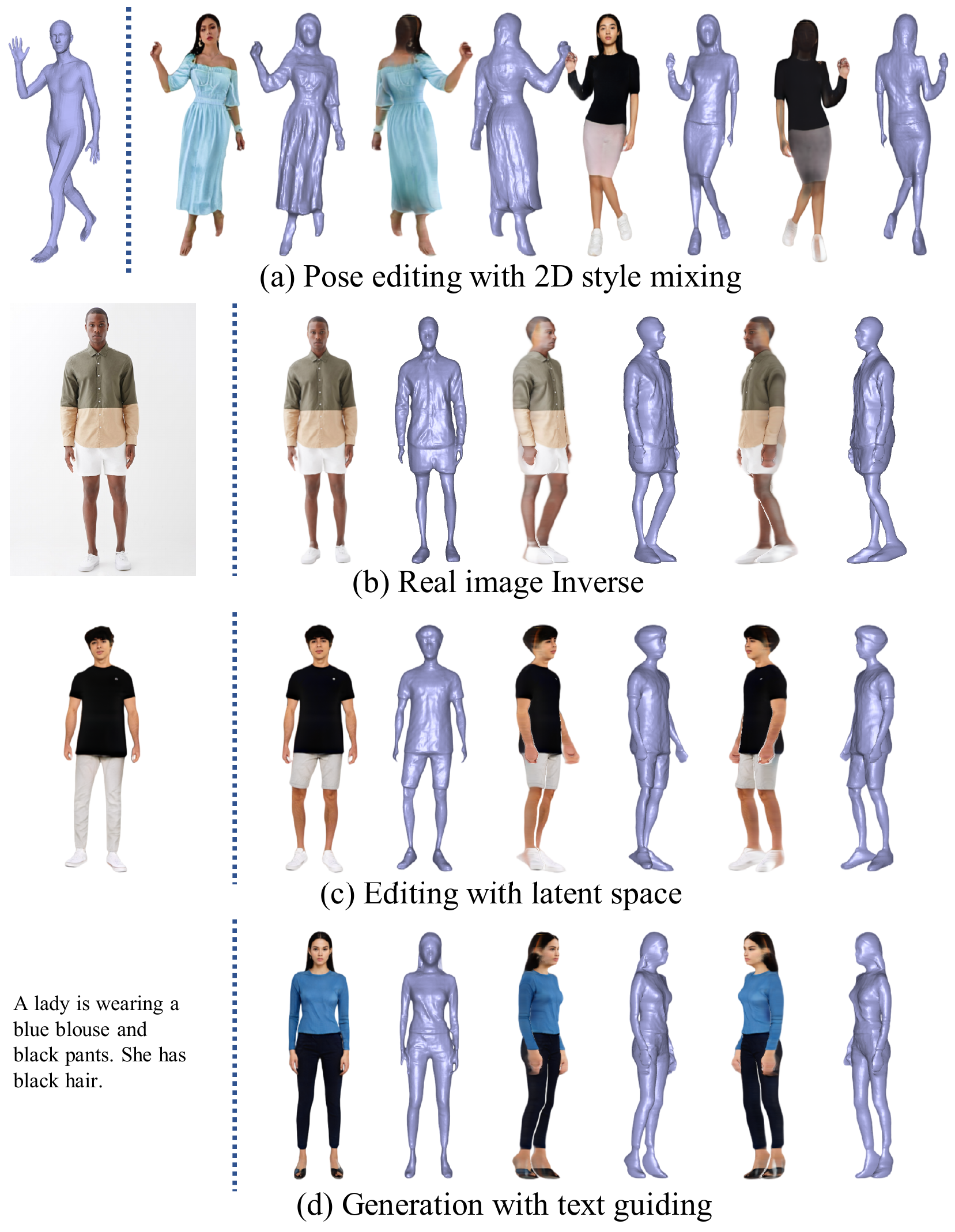} 
	\end{center} 
	\vspace{-20pt}
	\caption{Various 3D applications of HumanGen.} 
	\label{fig:app} 
	\vspace{-10pt}
\end{figure}

\myparagraph{Application.}
Our method is naturally compatible with existing 2D editing methods. As shown in Fig.~\ref{fig:app}(a), with style mixing on 2D, we are able to achieve 3D results with given pose. In Fig.~\ref{fig:app}(b), real images can be inverted to latent space to generate 3D results. In Fig.~\ref{fig:app}(c), by editing in latent space, the length change of upper or lower clothes can be seamlessly upgraded to 3D. In Fig.~\ref{fig:app}(d), 2D text-guided generation can be extended to 3D.

\subsection{Limitation and Discussion}

Although HumanGen achieves generating human with detailed geometry and $\ang{360}$ realistic rendering, it still has some limitations. First, the adopted view-balanced training set is essentially generated from a view-biased 2D generator, so the bias problem still potentially exists on uneven generation quality among different poses and views. It is still meaningful to have a real dataset with balanced view distribution and data of same identities. Second, with rich explicit priors in our method, it limits the out-of-domain generation ability of our method. Furthermore, our method does not support continuous skeletal pose control. It's promising to include motion prior into current framework for high-quality deformable generation, but requiring much more diverse training data. 

Though human eyes can still distinguish results of HumanGen, we care much about the ethical issue behind it. HumanGen should not be utilized to create any fake result of the real person and deceive people unfamiliar with this domain. Meanwhile, all our results are carefully chosen to ensure impartiality.

\section {Conclusion} 
We have presented a novel 3D human generation scheme with detailed geometry and $\ang{360}$ realistic free-view rendering. Our key idea is to introduce the concept of ``anchor image'' to aid the human generation using various human priors explicitly.
Our  hybrid feature representation efficiently bridges the latent space of HumanGen with the existing 2D generator.
Our geometry adapting scheme enables fine-grained details synthesis from 3D human reconstruction prior, while our two-stage blending scheme further encodes the rich texture information in the anchor image for appearance generation.
Our experimental results demonstrate the effectiveness of HumanGen for state-of-the-art 3D human generation. Various 3D applications of HumanGen further demonstrate its compatibility to existing off-the-shelf 2D editing toolbox based on latent disentanglement.
With the above unique ability, we believe that our approach is a critical step for high-quality 3D human generation, with various potential applications in VR/AR.

\end{CJK}

{\small
\bibliographystyle{ieee_fullname}
\bibliography{reference}
}

\begin{center}
\textbf{ }
\end{center}

\newpage
\section{Supplemental Material}
\subsection{Implementation Details}
We train our HumanGen using 4 NVIDIA A40 GPUs. We first train the geometry branch, which takes about 12 hours to converge. The trained implicit function $\mathbb{F}_{sdf}$ has hidden neurons of (273, 128, 32, 1). The loss weights are $\lambda_{mask}=1$, $\lambda_{3D\_SDF}=1$ and $\lambda_{eik}=0.1$. For the texture branch, we first train the base model for around 24 hours and then continue to train each model for another roughly 18 hours. For generating the tri-plane feature, we generate planes of shape $256\times256$ with channel size of $32$. We further add another two stylegan\cite{karras2020analyzing} synthesis blocks with up-scale equals to $1$ so as to extend the layer number of latent $w^{+}$ to $18$, which is compatible with the layer number of $w^{+}$ from Stylegan-human\cite{fu2022stylegan} generator. The loss weights are $\lambda_{2D\_front}=1$, $\lambda_{3D\_RGB}=8$, $\lambda_b=1\mathrm{e}{-2}$, $\lambda_{2D\_back}=8$ and the r1 regularization term of adversarial training (Eqn. 5) has weight $\lambda = 10$.

\subsection{Additional Evaluation}
\begin{figure}[t] 
    \vspace{-150pt}
	\begin{center} 
		\includegraphics[width=1\linewidth]{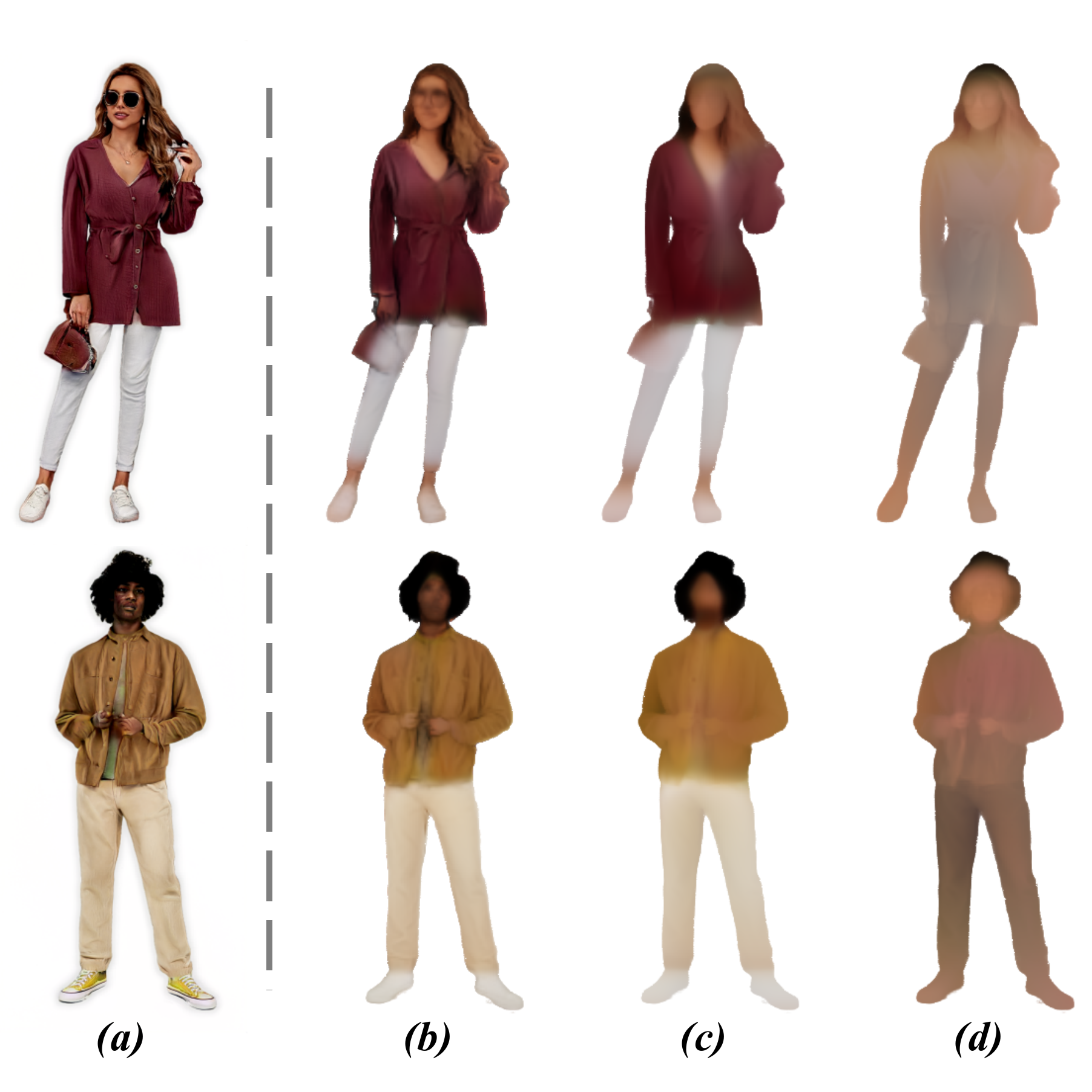} 
	\end{center} 
	\vspace{-20pt}
	\caption{Qualitative evaluation of texture fitting. (a) anchor image; (b) full; (c) w/o geometry embedding; (d) w/o $M$.}
	\label{fig:texture_fitting}
	\vspace{-10pt}
\end{figure}

\begin{figure*}[htbp] 
	\begin{center} 
		\includegraphics[width=0.99\linewidth]{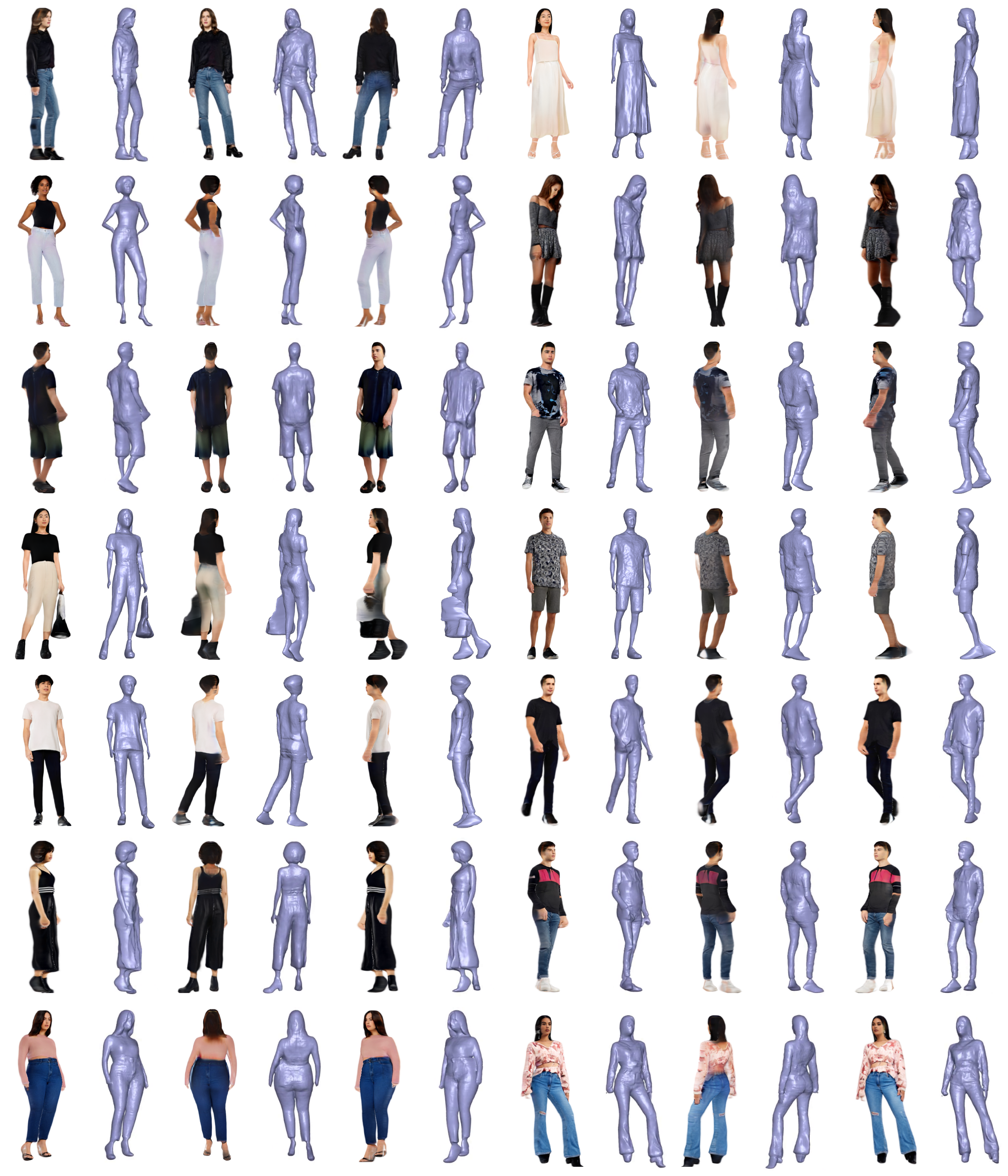} 
	\end{center} 
	\vspace{-20pt}
	\caption{More generation results using our HumanGen. Note that our approach enables photo-realistic human generation with detailed geometry and free-view rendering.}
	\label{fig:more_results}
	\vspace{-10pt}
\end{figure*}

We further conduct a texture fitting evaluation. Given the already-trained geometry branch, we only use anchor image and photometric loss to fit the frontal texture of given geometry. Let $M$ denotes the mapping network from Stylegan-human. As shown in Fig.\ref{fig:texture_fitting}, without the mapping network from Stylegan-human (\textbf{w/o $M$}), we train a newly-initialized mapping network to map the same noise $z$ to some latent and synthesize the tri-plane feature. However, it generates results that are all blurred and have closing color with each other because the newly-initialized mapping network fails to recover the original latent space of Stylegan-human. Without the geometry embedding feature given to the decoder (\textbf{w/o geometry embedding}), though the fitting texture can follow the color consistency with anchor image, it tends to have no details. With both $M$ and geometry embedding (\textbf{full}), the fitting results can maintain better consistency with anchor image while recovering some geometry details as well. The corresponding quantitative results are provided in Tab.\ref{tab:texture_fitting}.
However, the texture fitting results are still less-detailed. Therefore, we further add the blending scheme and GAN training to improve the effects. We provide more generation results of our complete model in Fig.\ref{fig:more_results}.
Note that our approach can generate photo-realistic results with detailed geometry and free-viewing ability. 

\begin{table}[t]
	\begin{center}
		\centering
		\caption{Quantitative evaluation of texture fitting.}
		\label{table:1}
		\resizebox{0.45\textwidth}{!}{
			\begin{tabular}{l|cccc}
				\hline
				Method      & PSNR$\uparrow$ & SSIM$\uparrow$ & LPIPS $\downarrow$  \\
				\hline
			
				w/o $M$       & 15.86            & 0.7835          & 0.2596      \\
                w/o geometry embedding      & 19.89            & 0.8236         & 0.2075      \\
				full        & \textbf{21.13}   & \textbf{0.8434} & \textbf{0.1803} \\
				\hline
			\end{tabular}
		}
            \label{tab:texture_fitting}
		\vspace{10pt}
	\end{center}
\end{table}

\end{document}